\documentclass{bmcart}

\usepackage[utf8]{inputenc} 
\usepackage{subfiles}
\usepackage{tabularx}
\usepackage{booktabs}
\usepackage{amsmath}
\usepackage{graphicx}
\usepackage{comment} 
\usepackage{graphicx}
\usepackage{float,tabularx} 
\usepackage{multirow}
\usepackage{graphicx}
\usepackage{tabularx}
\usepackage{siunitx}
\usepackage{amsmath}
\usepackage{amsthm}
\usepackage{lineno}

\usepackage{algorithm}
\usepackage{algpseudocode}

\DeclareMathOperator*{\agg}{AGG}

\usepackage{soul}
\usepackage{tikz}
\usetikzlibrary{calc}

\usepackage{colortbl}

\makeatletter
\newif\if@anonymize

\@anonymizefalse  

\if@anonymize
  \newcommand{\highlight@DoHighlight}{
    \fill [outer sep = -15pt, inner sep = 0pt, color=black]
          ($(begin highlight)+(0,8pt)$) rectangle ($(end highlight)+(0,-3pt)$) ;
  }

  \newcommand{\highlight@BeginHighlight}{
    \coordinate (begin highlight) at (0,0) ;
  }

  \newcommand{\highlight@EndHighlight}{
    \coordinate (end highlight) at (0,0) ;
  }

  \newdimen\highlight@previous
  \newdimen\highlight@current
  \newlength{\item@width}

  \DeclareRobustCommand*\anonymize{%
    \SOUL@setup
    \def\SOUL@preamble{%
      \begin{tikzpicture}[overlay, remember picture]
        \highlight@BeginHighlight
        \highlight@EndHighlight
      \end{tikzpicture}%
    }%
    \def\SOUL@postamble{%
      \begin{tikzpicture}[overlay, remember picture]
        \highlight@EndHighlight
        \highlight@DoHighlight
      \end{tikzpicture}%
    }%
    \def\SOUL@everyhyphen{%
      \discretionary{%
        \SOUL@setkern\SOUL@hyphkern
        \SOUL@sethyphenchar
        \tikz[overlay, remember picture] \highlight@EndHighlight ;%
      }{%
      }{%
        \SOUL@setkern\SOUL@charkern
      }%
    }%
    \def\SOUL@everyexhyphen##1{%
      \SOUL@setkern\SOUL@hyphkern
      \settowidth{\item@width}{##1}%
      \makebox[\item@width]{}%
      \discretionary{%
        \tikz[overlay, remember picture] \highlight@EndHighlight ;%
      }{%
      }{%
        \SOUL@setkern\SOUL@charkern
      }%
    }%
    \def\SOUL@everysyllable{%
      \begin{tikzpicture}[overlay, remember picture]
        \path let \p0 = (begin highlight), \p1 = (0,0) in \pgfextra
          \global\highlight@previous=\y0
          \global\highlight@current =\y1
        \endpgfextra (0,0) ;
        \ifdim\highlight@current < \highlight@previous
          \highlight@DoHighlight
          \highlight@BeginHighlight
        \fi
      \end{tikzpicture}%
      \settowidth{\item@width}{\the\SOUL@syllable}%
      \makebox[\item@width]{}%
      \tikz[overlay, remember picture] \highlight@EndHighlight ;%
    }%
    \SOUL@
  }
\else
  \newcommand{\anonymize}[1]{#1}
\fi
\makeatother

\startlocaldefs
\endlocaldefs

\begin{document}

\begin{frontmatter}

\begin{fmbox}
\dochead{Brief report}
\title{GCT-TTE: Graph Convolutional Transformer for Travel Time Estimation}

\author[
   noteref={n1},
   addressref={aff1},
   email={}
]{\inits{AM}\fnm{\anonymize{Vladimir}} \snm{\anonymize{Mashurov}}}
\author[
   noteref={n1},
   addressref={aff1},
   email={}
]{\inits{MK}\fnm{\anonymize{Vaagn}} \snm{\anonymize{Chopurian}}}
\author[
   noteref={n1},
   addressref={aff1,aff4},
   email={eighonet@gmail.com}
]{\inits{VP}\fnm{\anonymize{Vadim}} \snm{\anonymize{Porvatov}}}
\author[
   addressref={aff3},
   email={}
]{\inits{VT}\fnm{\anonymize{Arseny}} \snm{\anonymize{Ivanov}}}
\author[
   addressref={aff1,aff2},
    corref={aff2},      
   email={\anonymize{semenova.bnl@gmail.com}}
]{\inits{NS}\fnm{\anonymize{Natalia}} \snm{\anonymize{Semenova}}}

\address[id=aff1]{%
  \orgname{\anonymize{PJSC Sberbank}},
  \street{\anonymize{Vavilova Street}},
\postcode{\anonymize{117312}}
  \city{\anonymize{Moscow}},
  \cny{\anonymize{Russia}}
}
\address[id=aff4]{%
  \orgname{\anonymize{University of Amsterdam}},
  \street{\anonymize{Science Park 904}},
  \postcode{\anonymize{1098 XH}}
  \city{\anonymize{Amsterdam}},
  \cny{\anonymize{The Netherlands}}
}
\address[id=aff3]{
  \orgname{\anonymize{National University of Science and Technology "MISIS"}},
  \street{\anonymize{Lenin Avenue 4}},                  
  \postcode{\anonymize{119049}}                         
  \city{\anonymize{Moscow}},                            
  \cny{\anonymize{Russia}}                              
}
\address[id=aff2]{%
  \orgname{\anonymize{Artificial Intelligence Research Institute}},
  \street{\anonymize{Nizhniy Susalnyy Lane 5}},
  \postcode{\anonymize{105064}}
  \city{\anonymize{Moscow}},
  \cny{\anonymize{Russia}}
}

\begin{artnotes}
\note[id=n1]{Equal contribution}
\end{artnotes}

\end{fmbox}

\begin{abstractbox}

\begin{abstract} 
This paper introduces a new transformer-based model for the problem of travel time estimation. The key feature of the proposed GCT-TTE architecture is the utilization of different data modalities capturing different properties of an input path. Along with the extensive study regarding the model configuration, we implemented and evaluated a sufficient number of actual baselines for path-aware and path-blind settings. The conducted computational experiments have confirmed the viability of our pipeline, which outperformed state-of-the-art models on both considered datasets. Additionally, GCT-TTE was deployed as a web service accessible for further experiments with user-defined routes.   
\end{abstract}

\begin{keyword}
\kwd{machine learning}
\kwd{graph convolutional networks}
\kwd{transformers}
\kwd{geospatial data}
\kwd{travel time estimation}
\end{keyword}

\end{abstractbox}

\end{frontmatter}

\section*{Introduction}
Travel time estimation (TTE) is an actively developing branch of computational logistics that considers the prediction of potential time expenditures for specific types of trips~\cite{jenelius2013travel,wu2020estimate}. With the recent growth of urban environment complexity, such algorithms have become highly demanded both in commercial services and general traffic management~\cite{xuegang2010performance}. Following this line, better TTE decreases logistic costs for different kinds of delivery~\cite{Train_problem_TTE}, improves end-user experience for taxi services~\cite{Taxi_problem_TTE}, and ensures the quality of adaptive traffic control~\cite{technologies10010005}.

Despite the applied significance of travel time estimation, it still remains a challenging task in the case of ground vehicles. This situation arises from the influence of different patterns of road network topology, nonlinear traffic dynamics, changing weather conditions, and other types of unexpected temporal events. The majority of the currently established algorithms~\cite{wang2021graphtte,derrow2021eta} tend to utilize specific data modalities in order to capture complex spatio-temporal dependencies influencing the traffic flow. With the recent success of multimodal approaches in adjacent areas of travel demand prediction~\cite{traveldemand21} and journey planning~\cite{routing21}, fusing the features from different sources is expected to be the next step towards better performance in TTE. 

In this paper, we explored the predictive capabilities of TTE algorithms with different temporal encoders and proposed a new transformer-based model GCT-TTE. The main contributions of this study are the following:
\begin{enumerate}
\item In order to perform the experiments with the image modality, we extended the graph-based datasets for Abakan and Omsk~\cite{porvatov:2021} by the map patches (image modality) in accordance with the provided trajectories. Currently, the extended datasets are the only publicly available option for experiments with multimodal TTE algorithms.
\item In order to boost further research in the TTE area, we reimplemented and published the considered baselines in a unified format as well as corresponding weights and data preprocessing code. This contribution will enable the community to enhance evaluation quality in the future, as most of the TTE methods lack official implementations.
\item We proposed the GCT-TTE neural network for travel time estimation and extensively studied its generalization ability under various conditions. Obtained results allowed us to conclude that our pipeline achieved better performance regarding the baselines in terms of several metrics. Conducted experiments explicitly indicate that the performance of the transformer-based models is less prone to decrease (in the sense of the considered metrics) with the scaling of a road network size. This property remains crucial from an industrial perspective, as the classic recurrent models undergo considerably larger performance dropdowns.
\item For demonstration purposes, we deployed inference of the GCT-TTE model as the web application accessible for manual experiments. 
\end{enumerate}     

The web application is available at http://gctte.online and the code is published in the GitHub repository of the project\footnotemark[1]\footnotetext[1]{https://github.com/Eighonet/GCT-TTE}.

\section*{Related work}
Travel time estimation methods can be divided into two main types of approaches corresponding to the \textit{path-blind} and \textit{path-aware} estimation, Table~\ref{table:dataset}. The path-blind estimation refers to algorithms relying only on data about the start and end points of a route~\cite{wang2019simple}. The path-aware models use intermediate positions of a moving object represented in the form of GPS sequences~\cite{wang2014travel}, map patches~\cite{deepIST}, or a road subgraph~\cite{wang2021graphtte}. Despite the computational complexity increase, such approaches provide significantly better results, which justify the attention paid to them in the recent studies~\cite{zhang2018deeptravel,derrow2021eta,sun2021road}.

\begin{table}[t]
    \caption{Demonstration of utilized modalities in path-blind and path-aware models}
        
    \label{table:dataset}    
    \newcolumntype{P}[1]{>{\centering\arraybackslash}p{#1}}
    \renewcommand{\arraystretch}{1.6}
    
    \begin{minipage}[t]{.5\linewidth}

    \centering
    
    \begin{tabular}{|P{19.5mm}|P{8.1mm}|P{8.1mm}|P{8.1mm}|}
    \hline
    \multicolumn{4}{|c|}{Path-blind models}\\
    \hline
     \multicolumn{1}{|c|}{\multirow{2}{*}{Model}} & \multicolumn{3}{c|}{Modality} \\ 
     \cline{2-4} 
     & \multicolumn{1}{c|}{Graph} & \multicolumn{1}{c|}{Images} & \multicolumn{1}{c|}{GPS} \\
    
    \hline
    AVG & - & - & - \\
    LR & - & - & - \\
    MURAT & + & - & - \\
    DeepI2T & + & + & - \\
    \hline

    \end{tabular}
    \end{minipage}\hfill
    \begin{minipage}[t]{.5\linewidth}
    
    \centering    
    \begin{tabular}{|P{19.5mm}|P{8.1mm}|P{8.1mm}|P{8.1mm}|}
    \hline
    \multicolumn{4}{|c|}{Path-aware models}\\
    \hline
     \multicolumn{1}{|c|}{\multirow{2}{*}{Model}} & \multicolumn{3}{c|}{Modality} \\ 
     \cline{2-4} 
     & \multicolumn{1}{c|}{Graph} & \multicolumn{1}{c|}{Images} & \multicolumn{1}{c|}{GPS} \\
    
    \hline
    WDR & + & - & - \\
    DeepIST & - & + & - \\
    DeepTTE & - & - & + \\
    DeepI2T & + & + & - \\
    \hline
    \end{tabular}
    \end{minipage}
\end{table}

One of the earliest path-aware models was the wide-deep-recurrent (WDR) architecture~\cite{wang2018wdr}, which mostly inherited the concept of joint learning from recommender systems~\cite{cheng2016wide}. In further studies, this approach was extended regarding the usage of different data modalities. In particular, the DeepIST~\cite{deepIST} model utilizes rectangular fragments of a general reference map corresponding to elements of a route GPS sequence. Extracted images are fed into a convolutional neural network (CNN) that captures spatial patterns of depicted infrastructure. These feature representations are further concatenated into the matrix processed by the long short-term memory (LSTM) layer~\cite{hochreiter1997long}.

In contrast with the other approaches, DeepTTE~\cite{DeepTTE} is designed to operate directly on GPS coordinates via geospatial convolutions paired with a recurrent neural network. The first part of this pipeline transforms raw GPS sequences into a series of feature maps capturing the local spatial correlation between consecutive coordinates. The final block learns the temporal relations of obtained feature maps and produces predictions for the entire route along with its separate segments.

The concept of modality fusing was first introduced in TTE as a part of the DeepI2T~\cite{deepI2T} model. This architecture uses large-scale information network embedding~\cite{line} to produce grid representations and 3-layer CNN with pooling for image processing. As well as DeppTTE, DeepI2T includes the segment-based prediction component implemented in the form of residual blocks on the top of the Bi-LSTM encoder.

In addition to extensively studied recurrent TTE methods, it is also important to mention recently emerged transformer models~\cite{liu2022mctte,semenova2022logistics}. Despite the limited comparison with classic LSTM-based methods, they have already demonstrated promising prediction quality, preserving the potential for further major improvements~\cite{ttpnet22,mtte21}. As most of the transformer models lack a comprehensive evaluation, we intend to explore GCT-TTE performance with respect to a sufficient number of state-of-the-art solutions to reveal its capabilities explicitly.

\section*{Preliminaries}
\begin{figure*}[t!]
  \centering
  \includegraphics[width=348pt]{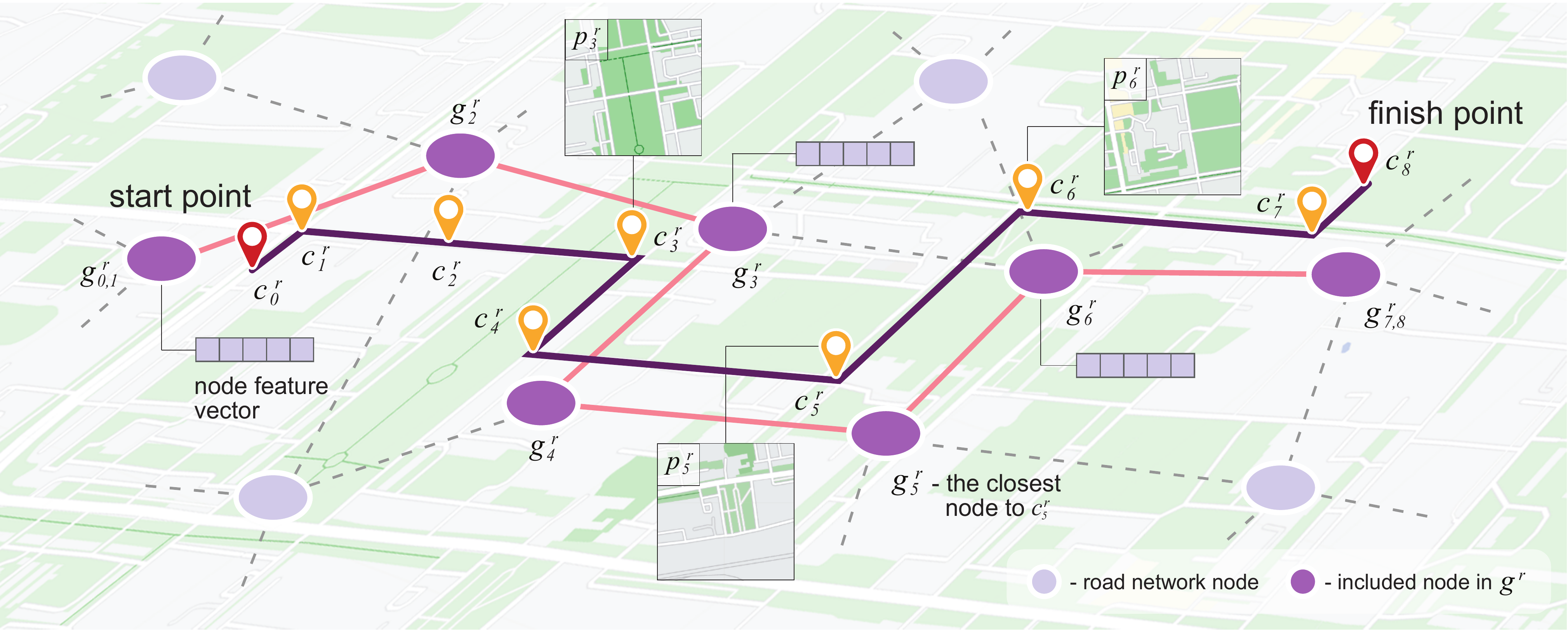}
  \caption{Example of data modalities of an arbitrary route $r$: for each GPS coordinate $c^r_i$ from $c^r$, there is a corresponding node $g^r_i$ with associated features and map patch $p^r_i$.}
  \label{fig:data_mod}
\end{figure*}

In this section, we introduce the main concepts required to operate with the proposed model, Figure~\ref{fig:data_mod}.

\textbf{Route}. A route $r$ is defined as the set $\{c^r, a^r, t^r\}$, where $c^r$ is the sequence of GPS coordinates of a moving object, $a^r$ is the vector of temporal and weather data, $t^r$ is the travel time.

As the \textit{image modality} $p^r$ of a route $r$, we use geographical map patches corresponding to each coordinate $c^r_i \in c^r$. Each image has a fixed size $ 256 \times 256 \times 3$ across all of the GPS sequences in a specific dataset. 

\textbf{Road network}. Road network is represented in the form of graph $G = \{V, E, X\}$, where $V = \{v_1,\;...\;,v_n\}$ is the set of nodes corresponding to the segments of city roads, $E = \{(v_i, v_j) \; | \; v_i \rightarrow v_j\}$ is the set of edges between connected nodes $v_i, v_j \in V$, $X: n \times m \rightarrow \mathbf{R}$ is a feature matrix of nodes describing properties of the roads' segments (additional information regarding available graph features is provided in Supplementary 1).

 Description of a route $r$ can be further extended by the \textit{graph modality} $g^{r} = \{v_k\:|\:k=argmin_{j} \, \rho(c^r_i, v_j)\}^{|c^r|}_{i=1}$, where $\rho(c^r_i, v_j)$ is the minimum Euclidean distance between coordinates associated with $v_j$ and $c_i^r$. Following the same concept as in the case of $p^r$, the graph modality represents a sequence of nodes and their features aggregated with respect to the initial GPS coordinates $c^r$.    
 
\textbf{Travel time estimation}. For each entry $r$, it is required to estimate the travel time $t^r$ using the elements of feature description $\{c^r, p^r, g^r, a^r\}$.

\section*{Data}
\begin{figure*}[t!]
  \centering
  \includegraphics[width=348pt]{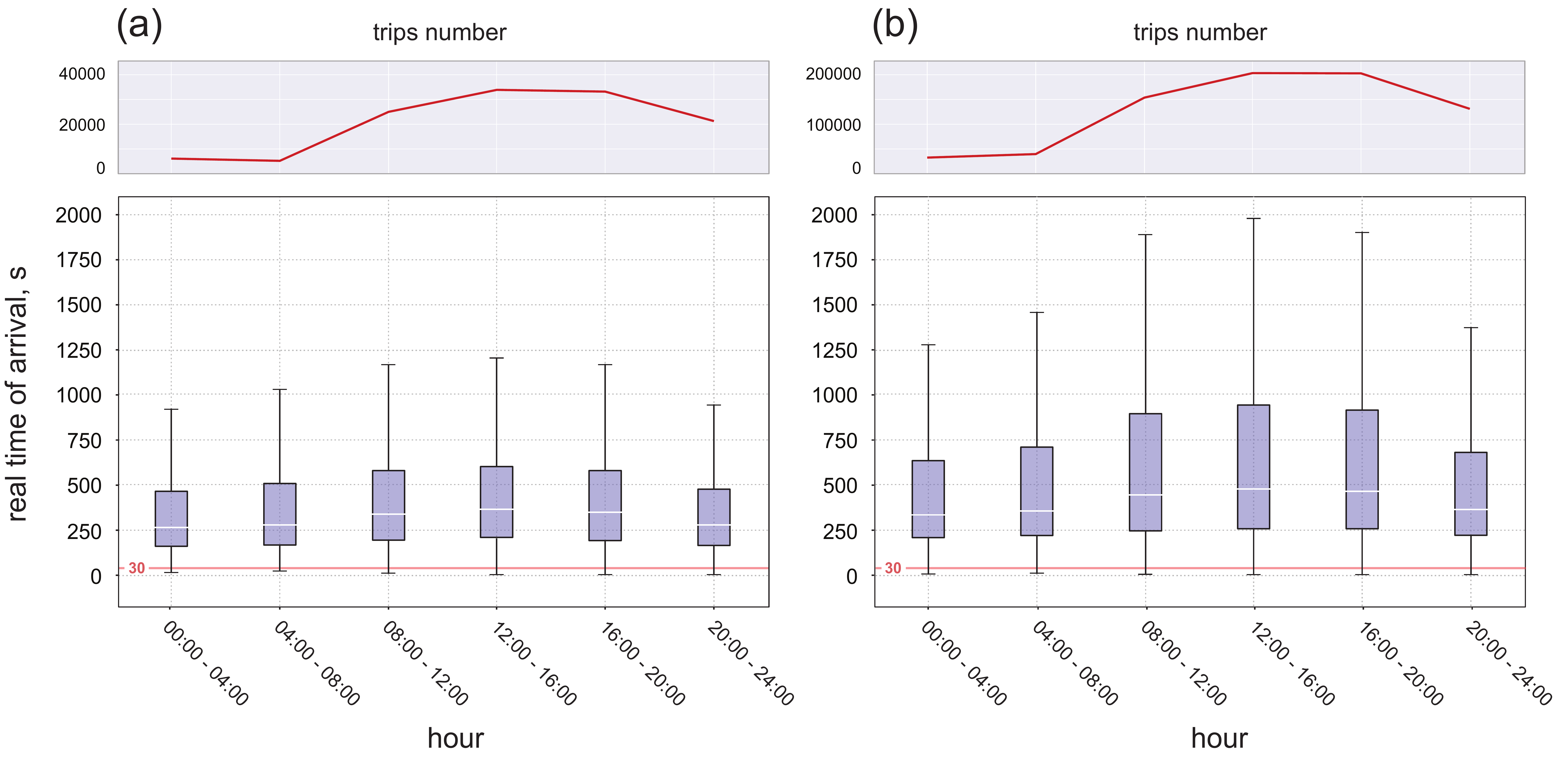}
  \caption{Cumulative frequencies of car activity and distribution of trips duration for Abakan (a) and Omsk (b) in the four hours interval.}
  \label{fig:distr_stats}
\end{figure*}

We explored the predictive performance of the algorithm on two real-world datasets collected during the period from December 1, 2020 to December 31, 2020 in Abakan (112.4 square kilometers) and Omsk (577.9 square kilometers). Each dataset consists of a road graph and associated routes, Table~\ref{table:data}. In the preprocessing stage, we excluded trips that lasted less than 30 seconds (273, 0.22\% for Abakan; 1194, 0.15\% for Omsk) along with the ones that took more than 50 minutes (223, 0.18\% for Abakan; 3681, 0.47\% for Omsk). The distributional statistics of both datasets are depicted in Figure~\ref{fig:distr_stats}.

\begin{table}[t!]

    \caption{Description of the Abakan and Omsk datasets.
    \label{table:data}
    }
    
    \newcolumntype{P}[1]{>{\centering\arraybackslash}p{#1}}
    \renewcommand{\arraystretch}{1.6}

    \begin{minipage}[t]{.5\linewidth}
    
    \centering
    
    \begin{tabular}{|P{21mm}|P{14.0mm}|P{14.0mm}|}
    
    \hline
    \multicolumn{3}{|c|}{Road network}\\
    \hline
    \multicolumn{1}{|c|}{Property$\:\backslash\:$City} & \multicolumn{1}{c|}{Abakan} & \multicolumn{1}{c|}{Omsk} \\  
    \hline
     Nodes & $65\,524$ &	$231\,688$ \\
     Edges & $340\,012$ & $1\,149\,492$ \\
     Clustering & 0.5278 & 0.53 \\
     Usage median & 12 & 8 \\
    \hline

    \end{tabular}
    \end{minipage}\hfill
    \begin{minipage}[t]{.5\linewidth}
    
    \centering
    
    \begin{tabular}{|P{21mm}|P{14.0mm}|P{14.0mm}|}
    
    \hline
    \multicolumn{3}{|c|}{Trips}\\
    \hline
    \multicolumn{1}{|c|}{Property$\:\backslash\:$City} & \multicolumn{1}{c|}{Abakan} & \multicolumn{1}{c|}{Omsk} \\  
    \hline
     Trips number & $121\,557$ & $767\,343$ \\
     Coverage & 53.3\% & 49.5\% \\
     Average time & 427 sec & 608 sec \\
     Average length & 3604 m & 4216 m \\
    \hline

    \end{tabular}
    \end{minipage}
\end{table}

Since initial versions of Abakan and Omsk datasets did not have any relevant input data for image-based models, we extended their road graphs with the map patches parsed from Open Street Map (OSM)\footnotemark[2]\footnotetext[2]{https://www.openstreetmap.org}. The parsing algorithm extracted patches from the OSM tile server URLs in accordance with the following request template: http://a.tile.openstreetmap.org/
\{zoom\}/\{longitude\}/\{latitude\}.png. In order to utilize obtained data together with initial graphs, it was extended by mapping tables including the closest node id for each patch. The applied proximity measure was based on the Euclidean distance between location of image centroids and geographical coordinates of graph vertexes. Due to the limitations of the API throughput, the procedure of image extraction was distributed between several machines with a total execution time exceeding 1 week.

The provided extension consists of images dated July 2022: due to the absence of significant changes in the road network topology since 2020, image modality for Abakan and Omsk remains actual with respect to the original graph-based data. The content of the patches includes a full range of geographic objects useful for travel time estimation (e.g., road networks, landscape groups, buildings and associated infrastructural objects) and covers all of the routes provided in the initial datasets.

Depending on the requirements of the considered learning model, image datasets had to be organized regarding the fixed grid partitions or centered around the elements of GPS sequences. In the first case, a geographical map of a city was divided into equal disjoint patches, which were further mapped with the GPS data in accordance with the presence of coordinates in a specific partition. Trajectory-based approach to dataset construction does not require the disjoint property of images and relies on the extraction of patches with the center in the specified coordinate, Algorithm~\ref{alg:route_prep} (\textit{collect} and \textit{split} functions can be accessed in Supplementary 2-3). The obtained grid-based image dataset consists of $96 \, 101$ instances for Abakan and $838 \, 865$ for Omsk while the trajectory-based dataset has $544 \, 502$ and $3 \, 376 \, 294$ images correspondingly.

One of the crucial features of the considered datasets is the absence of traffic flow properties. The availability of such data is directly related to the specialized tracking systems (based on loop detectors or observation cameras), which are not presented in the majority of cities. In order to make the GCT-TTE suitable for the greatest number of urban environments, we decided not to limit the study by the rarely accessible data.

\section*{Method}
\begin{figure*}[t!]
  \centering
  \includegraphics[width=348pt]{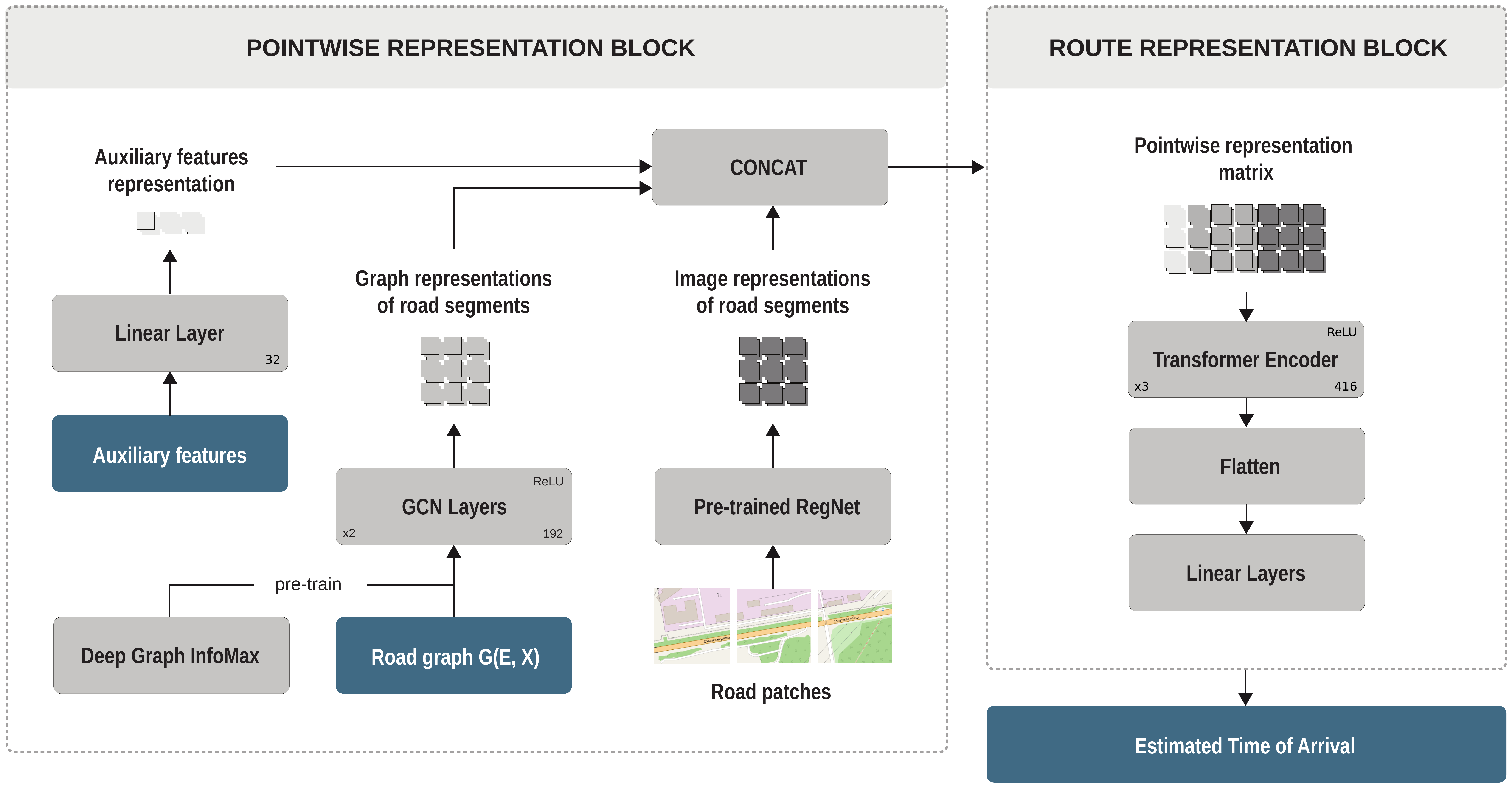}
  \caption{Demonstration of the GCT-TTE pipeline: feature extraction algorithms applied to considered modalities and extended by transformer encoder capturing the concatenated sequence of embeddings.}
  \label{fig:gct-tte}
\end{figure*}

In this section, we provide an extensive description of the GCT-TTE main components: pointwise and sequence representation blocks, Figure~\ref{fig:gct-tte}.  

\subsection*{Patches encoder}

In order to extract features from the image modality, we utilized the RegNetY~\cite{RegNet} architecture from the SEER model family. The key component of this architecture is the convolutional recurrent neural network (ConvRNN) which controls the spatio-temporal information flow between building blocks of the neural network.

Each RegNetY block consists of three operators. The initial convolution layer of $t$'th block processes the input tensor $X^t_1$ and returns the feature map $X^t_2$. Next, the obtained representation $X^t_2$ is fed to ConvRNN:
\begin{equation}
    H^t = \mathrm{tanh}(\mathrm{C_x}(X^t_2) + 
    \mathrm{C_h}(H^{t-1}) + b_h),
\end{equation}
where $H^{t-1}$ is the hidden state of the previous RegNetY block, $b_h$ is a bias tensor, $\mathrm{C_x}$ and $\mathrm{C_h}$ correspond to convolutional layers. In the following stage, $X^t_2$ and $H^t$ are fed as input to the last convolution layer, which is further extended by residual connection.

As the SEER models are capable of producing robust features that are well-suited for out-of-distribution generalization~\cite{pretrainRegNet}, we pre-trained RegNetY with the following autoencoder loss:
\begin{equation}
    \mathcal{L}(W \times RegNet(X), \, f(X)) \rightarrow 0,
\end{equation}
where $\mathcal{L}$ is the binary cross-entropy function, $f$ is an image flattening operator, and $W$ is the projection matrix of learning parameters that maps model output to the flattened image.

\begin{algorithm}[t]

\caption{\small{Trajectory-based route preparation algorithm for image modality}}
\label{alg:route_prep}

\begin{algorithmic}[1]

\State $result = [ \; ]$ \Comment{$[\;]$ -- empty list}

\For{$path$ in $paths$} \Comment{For each path}
\State $route\_centers = [ \; ]$ 
\State $i \gets 0$

    \If{len(path) == 1} \Comment{If path contents only one node}
        \State $result \gets path$
        \State continue
    \EndIf

    \While{$i + 1 < len(path)$} \Comment{len() -- length}
        \State $centroids \gets [ \; ]$
        \State $k \gets 1$
        \If{($\rho(path[i], path[i+1]) < 200$)}
            \State $center, k \gets collect(path, i)$
            \State $route\_centers \gets center$
            \State i = i + k
            \State continue
        \EndIf
        \If{$\rho(path[i], path[i+1]) >= 200$}
            \For{$point$ in $split(path[i], path[i+1])$}
                \State $route\_centers \gets point$
                \EndFor
        \EndIf
        \State $i = i + 1$
    \EndWhile
    \State $result \gets route\_centers$
\EndFor
\end{algorithmic}
\end{algorithm}

\subsection*{Auxiliary encoder}
Along with the map patches and graph elements, we apply additional features $a^r$ corresponding to the temporal and weather data (e.g., trip hour, type of day, precipitation). The GCT-TTE model processes this part of the input with the help of a trivial linear layer: 
\begin{equation}
A^{r} = W a^r,
\end{equation}
where $W$ is a matrix of learning parameters.

\subsection*{Graph encoder}
\label{sec:pwrb}

The graph data is handled with the help of the graph convolutional layers defined as follows:
\begin{equation}
h_u^{(k)} = \mathrm{ReLU}\left( W^{(k)} \agg_{v \in \mathcal{N}(u)}\,(\frac{h^{(k-1)}_v}{||{\mathcal{N}_{uv}}||})\right),
\end{equation}
where $h_u^{(k)}$ is a $k$-hop embedding~\cite{GCN} of $u \in V$, $h_u^{(0)} = x_u$, $W^{(k)}$ is a matrix of learning parameters of $k$'th convolutional layer, $\mathcal{N}(u)$ is a set of neighbour nodes of $u$, $\agg_{v \in \mathcal{N}(u)}$ is a sum aggregarion function, and $||{\mathcal{N}_{uv}}|| = \sqrt{|\mathcal{N}(u)||\mathcal{N}(v)|}$. 

To accelerate the convergence of the GCT-TTE model, we pre-trained the weights of the graph convolutions by the Deep Graph InfoMax algorithm~\cite{velickovic2018deep}. This approach optimizes the loss function that allows learning the difference between initial and corrupted embeddings of nodes: 
\begin{equation}
\mathcal{L} = \frac{1}{N+M} \Big(\sum_{i=1}^N {E}_{\mathcal{G}} \Big[log(D(h_u, h_\mathcal{G}))\Big] + \sum_{j=1}^M\ {E}_{\tilde{\mathcal{G}}}\left[log(1-D(\tilde{h}_u, h_\mathcal{G}))\right]\Big),
\end{equation}
where $h_u$ is an embedding of node $u$ based on the initial graph $\mathcal{G}$, $\tilde{h}_u$ is an embedding of a node $u$ from the corrupted version $\tilde{\mathcal{G}}$ of the graph $\mathcal{G}$, $D$ corresponds to the discriminator function.
\\

The final output of the pointwise block constitutes a concatenation of the weighted representations and auxiliary data for each route $r$ with $k$ segments:
\begin{equation}
P_r = \mathrm{CONCAT}(\alpha \cdot H^r, (1 - \alpha) \cdot I^r, \; \beta \cdot A^r),
\end{equation}
where $H^r$ is the matrix of size $k \times e_g$ of graph-based segment embeddings, $I^r$ is the matrix of size $k \times e_i$ obtained from a flattened RegNet output, $\alpha$, $(1 - \alpha)$, and $\beta$ correspond to the weight coefficients of specific modalitites.     

\subsection*{Sequence representation block}

To extract sequential features from the output of the pointwise representation block, it is fed to transformer encoder~\cite{AttentionIsAllYouNeed}. The encoder consists of two attention layers with a residual connection followed by a normalization operator. The multi-head attention coefficients are defined as follows:
\begin{equation}
    \alpha^{(h)}_{i, j} = \mathrm{softmax}_{\,w_j}\left(\frac{\langle W^T_{h, q}x_i, W^T_{h, k}x_j \rangle}{\sqrt{d_k}}\right),
\end{equation}
where $x_i, x_j \in P_r$, $h$ is an attention head,  $d_k$ is a scale coefficient, $W^T_{h, q}$ and $ W^T_{h, k}$ are query and key weight matrices, $w_j$ is a vector of softmax learning parameters. The output of the attention layer will be:
\begin{equation}
    u_i = \mathrm{LayerNorm}\left(x_i + \sum_{h=1}^H W^T_{c, h} \sum_{j=1}^n \alpha_{i, j}^{(h)} W^T_{h, v}x_j \right),
\end{equation}
where $W^T_{h, v}$ is value weight matrix, $H$ is a number of attention heads.

The final part of the sequence representation block corresponds to the flattening operator and several linear layers with the ReLU activation, which predict the travel time of a route.

\section*{Results}
In this section, we reveal the parameter dependencies of the model and compare the results of the considered baselines.

\subsection*{Experimental setup}

The experiments were conducted on 16 GPU Tesla V100. For the GCT-TTE training, Adam optimizer~\cite{kingma2014adam} was chosen with a learning rate $5\cdot10^{-5}$ and batch size of 16. For better convergence, we apply the scheduler with patience equal to 10 epochs and 0.1 scaling factor. The training time for the final configuration of the GCT-TTE model is 6 hours in the case of Abakan and 30 for Omsk.     

The established values of quality metrics were obtained from the 5-fold cross-validation procedure. As the measures of the model performance, we use mean absolute error (MAE), rooted mean squared error (RMSE), and 10$\%$ satisfaction rate (SR). Additionally, we compute mean absolute percentage error (MAPE) as it is frequently applied in related studies.

\subsection*{Models comparison and evaluation}
The results regarding path-blind evaluation are depicted in Table~3. Neighbor average (AVG) and linear regression (LR) demonstrated the worst results among the trivial baselines as long as gradient boosted decision trees (GBDT) explicitly outperformed more complex models in the case of the largest city. The MURAT model achieved the best score for Abakan in terms of MAE and RMSE, while GCT-TTE has the minimum MAPE among all of the considered architectures. 

Demonstrated variability of metric values makes the identification of the best model rather a hard task for a path-blind setting. The simplest models are still capable to be competitive regarding such architectures as MURAT, which was expected to perform tangibly better on both considered datasets. The results regarding GCT-TTE can be partially explained by its structure as it was not initially designed for a path-blind evaluation. 

As can be seen in Table~4, the proposed solution outperformed baselines in terms of the RMSE value, which proves the rigidity of GCT-TTE towards large errors prevention. The comparison of MAE and RMSE for considered methods has shown a minimal gap between these metrics in the case of GCT-TTE for both cities, signifying the efficiency of the technique with respect to dataset size. Overall, the results have confirmed that GCT-TTE appeared to be a more reliable approach than the LSTM-based models: while MAPE remains approximately the same across top-performing architectures, GCT-TTE achieves significantly better MAE and RMSE values. Conducted computational experiments also indicated that DeepI2T and WDR have intrinsic problems with the convergence, while GCT-TTE demonstrates smoother training dynamics.
 
\begin{table*}[!t]

    \renewcommand{\arraystretch}{1.6}

    \centering

    \newcolumntype{Y}{>{\centering\arraybackslash}X}

    \sisetup{detect-weight,mode=text}
    \renewrobustcmd{\bfseries}{\fontseries{b}\selectfont}
    \renewrobustcmd{\boldmath}{}
    \newrobustcmd{\B}{\bfseries}
    \begin{minipage}[t]{1\linewidth}
    \label{table:results_path_blind}
    \caption{Path-blind models comparison}
     \begin{tabularx}{360px}{|XX|Y|Y|Y|Y|Y|Y|Y|Y|}
    \hline
        \multicolumn{2}{|c|}{} & \multicolumn{4}{c|}{Abakan} & \multicolumn{4}{c|}{Omsk} \\ \hline
       \multicolumn{2}{|c|}{\textit{$\;$Baseline $\backslash$Metric$\;$}} & MAE & RMSE & MAPE & SR & MAE & RMSE & MAPE & SR \\ \hline
        \multicolumn{2}{|c|}{AVG} & 322.77 & 477.61 & 0.761 & 0.018 & 439.05 & 628.75 & 0.741 & 0.012 \\ \hline
        \multicolumn{2}{|c|}{LR} & 262.33 & 456.63 & 1.169 & 9.527 & 416.81 & 593.01 & 1.399 & 7.187 \\ \hline
        \multicolumn{2}{|c|}{GBDT} & 245.77 & 433.91 & 1.106 & 10.28 & \textbf{209.99} & \textbf{372.11} & 0.656 & \textbf{17.72} \\ \hline \hline
        \multicolumn{2}{|c|}{MURAT} & \textbf{182.97} & \textbf{282.15} & 0.685 & 10.77 & 285.72 & 444.74 & 0.856 & 9.997 \\ \hline
        \hline
        \multicolumn{2}{|c|}{GCT-TTE} & 221.71 & 337.59 & \textbf{0.505} & \textbf{11.12} & 376.74 &	590.93 & \textbf{0.5486} & 8.99 \\ \hline
    \end{tabularx}
    \end{minipage}
    \vspace{2px}
    
    \begin{minipage}[t]{1\linewidth}
    \label{table:results_path_aware}
    \caption{Path-aware models comparison}
     \begin{tabularx}{360px}{|XX|Y|Y|Y|Y|Y|Y|Y|Y|}
    \hline
        \multicolumn{2}{|c|}{} & \multicolumn{4}{c|}{Abakan} & \multicolumn{4}{c|}{Omsk} \\ \hline
        \multicolumn{2}{|c|}{\textit{$\;$Baseline$\backslash$Metric$\;$}} & MAE & RMSE & MAPE & SR & MAE & RMSE & MAPE & SR \\ \hline
        \multicolumn{2}{|c|}{DeepIST} & 153.88 & 241.29 & 0.3905 & 18.08 & 256.50 & 415.16 & 0.6361 & 14.39 \\ \hline
        \multicolumn{2}{|c|}{DeepTTE} & 111.03 & 174.56 & 0.2165 & 31.45 & 179.07 & 296.98 & \textbf{0.1898} & 34.03 \\ \hline
        \multicolumn{2}{|c|}{GridLSTM} & 100.27 & 206.91 & 0.2202 & 30.74 & 135.74 & 257.18 & 0.2120 & 31.21 \\ \hline
        \multicolumn{2}{|c|}{DeepI2T} & 97.99 & 201.33 & \textbf{0.2128} & 31.34 & 136.66 & 260.90 & 0.2124 & 31.23 \\ \hline 
        \multicolumn{2}{|c|}{WDR} & 97.22 & 190.09 & 0.2162 & \textbf{31.98} & 131.57  & 269.00 & 0.2039 & 33.34 \\ \hline
        \hline
        \multicolumn{2}{|c|}{GCT-TTE} & \textbf{92.26} & \textbf{147.89} & 0.2262 & 30.46 & \textbf{107.97} & \textbf{169.15} & 0.1961 & \textbf{35.17} \\ \hline
    \end{tabularx}
    \end{minipage}
\end{table*}

\begin{figure*}[t!]
  \centering
  \includegraphics[width=348pt]{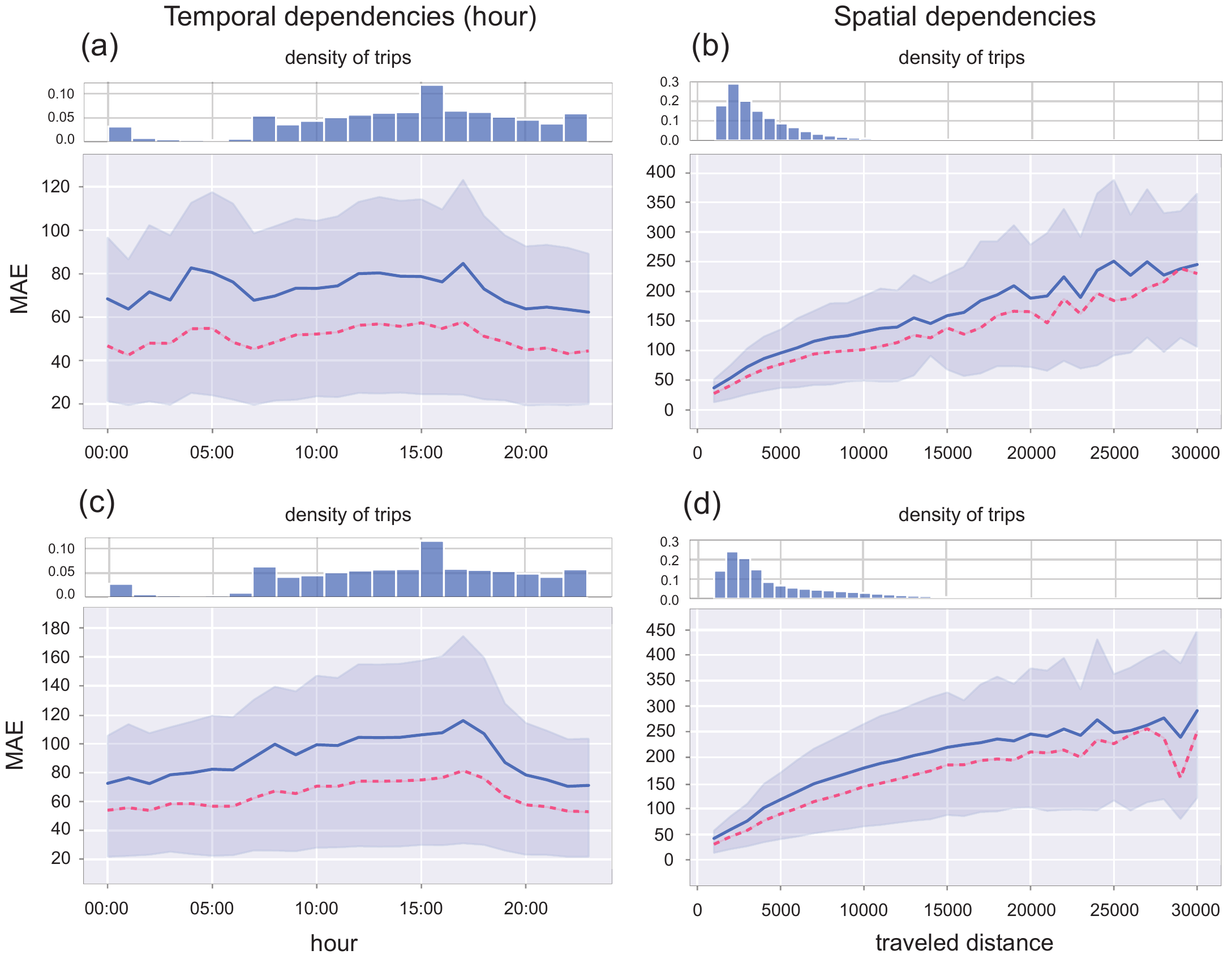}
   \caption{Spatial and temporal (hour) dependencies across the different groups of test entries for Abakan (a, b) and Omsk (c, d): blue and red lines depict mean and median values of MAE, borders of filled area correspond to Q1 and Q3 quartiles of a MAE distribution.}
    \label{fig:spatial_and_temporal_relations}
\end{figure*}

\begin{figure*}[ht!]
  \centering
  \includegraphics[width=348pt]{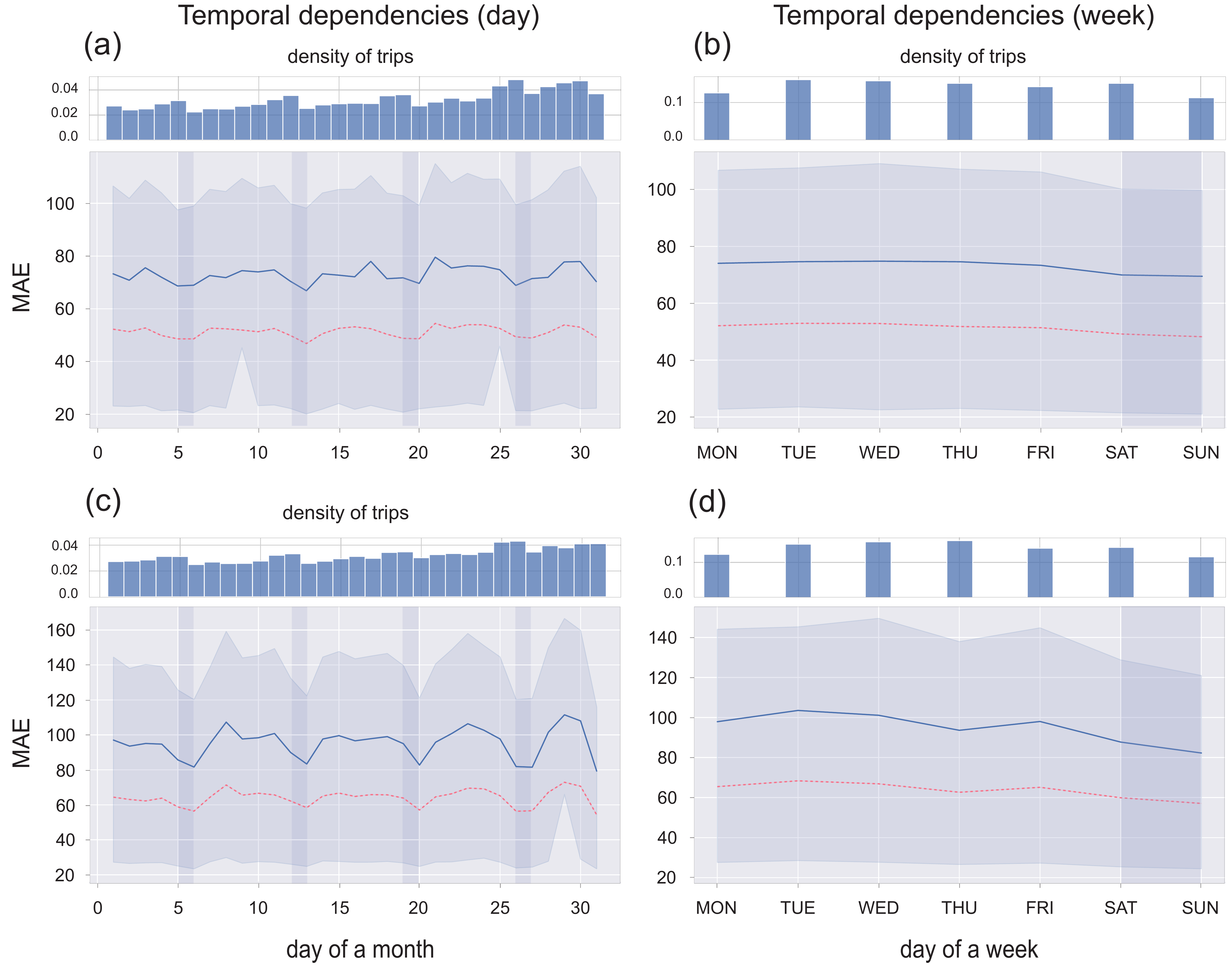}
   \caption{Temporal (day and week) dependencies across the different groups of test entries for Abakan (a, b) and Omsk (c, d): blue and red lines depict mean and median values of MAE, borders of filled area correspond to Q1 and Q3 quartiles of a MAE distribution. The weekends are represented by the vertical areas filled with a darker colour.}
   \label{fig:day_week_relations}
\end{figure*}

\subsection*{Performance analysis}
In the case of both datasets, dependencies between the travelled distance and obtained MAE on the corresponding trips reveal similar dynamics: as the path length increases, the error rate continues to grow, Figure~\ref{fig:spatial_and_temporal_relations}(b, d). The prediction variance is inversely proportional to the number of routes in a particular length interval except for the small percentage of the shortest routes. The main difference between the MAE curves is reflected in the higher magnitudes of performance fluctuations in Abakan compared to Omsk.

The temporal dynamics of GCT-TTE errors exhibit rich nonlinear properties during a 24-hour period. The shape of the error curves demonstrates that our model tends to accumulate a majority of errors in the period between 16:00 and 18:00, Figure~\ref{fig:spatial_and_temporal_relations}(a, c). This time interval corresponds to the end of the working day, which has a crucial impact on the traffic flow foreseeability. 

Despite the mentioned performance outlier, the general behaviour of temporal dependencies allows concluding that GCT-TTE successfully captures the factors influencing the target value in the daytime. With the growing sparsity of data during night hours, it is still capable of producing relevant predictions for Omsk. In the case of Abakan, the GCT-TTE performance drop can be associated with a substantial reduction in intercity trips number (which emerged to be an easier target for the model).

Focusing on higher levels of seasonality, day- and week-based temporal dependencies of error demonstrate explicit periodical behaviour, Figure~\ref{fig:day_week_relations}. The GCT-TTE model performs better at the end of the week for both considered cities, with a pronounced error decrease in the case of Omsk. In contrast, the middle of the week (i.e. Wednesday for Abakan and Tuesday for Omsk) is the most challenging period, which has averagely 12.48\% higher MAE compared to Saturday and Sunday. 

\begin{figure*}[t!]
  \centering
  \includegraphics[width=348pt]{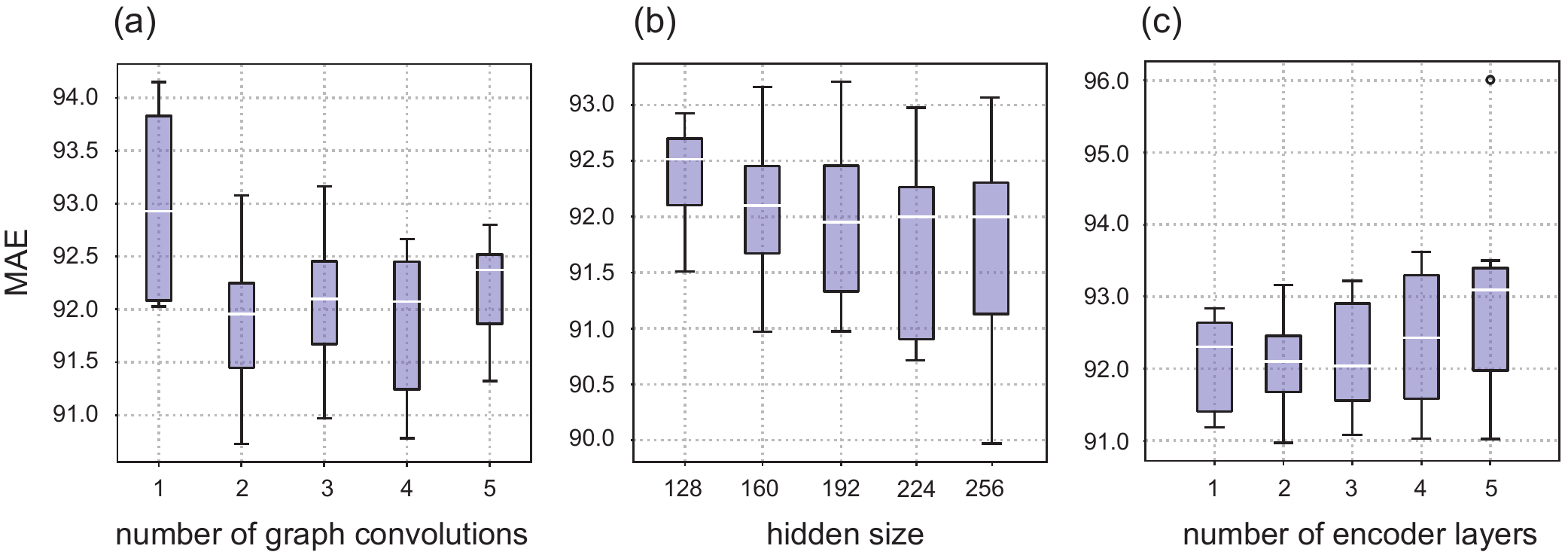}
   \caption{Parametric dependencies of GCT-TTE performance for Abakan: number of graph convolutions (a), hidden size of graph convolutions (b), and number of transformer encoder layers (c).}
    \label{fig:sensitivity}
\end{figure*}

\subsection*{Sensitivity analysis}
In order to achieve better prediction quality, we extensively studied the dependencies between GCT-TTE parameters and model performance in the sense of the MAE metric. The best value for modality coefficient $\alpha$ was 0.9, which reflects the significant contribution of graph data towards error reduction. For the final model, we utilized 2 graph convolutional layers with hidden size 192, Figure~\ref{fig:sensitivity}(a, b). The lack of aggregation depth can significantly reduce the performance of GCT-TTE, while the excessive number of layers has a less expressive negative impact on MAE. A similar situation can be observed in the case of the hidden size, which is getting close to a plateau after reaching a certain threshold value.  

Along with the graph convolutions, we explored the configuration of the sequence representation part of GCT-TTE. Since the transformer block remains its main component, the computational experiments were focused on the influence of encoder depth on quality metrics, Figure~3(c). As it can be derived from the U-shaped dependency, the best number of attention layers is 3. 

\section*{Demonstration}
\begin{figure*}[!t]
  \centering
  
  \includegraphics[width=348pt]{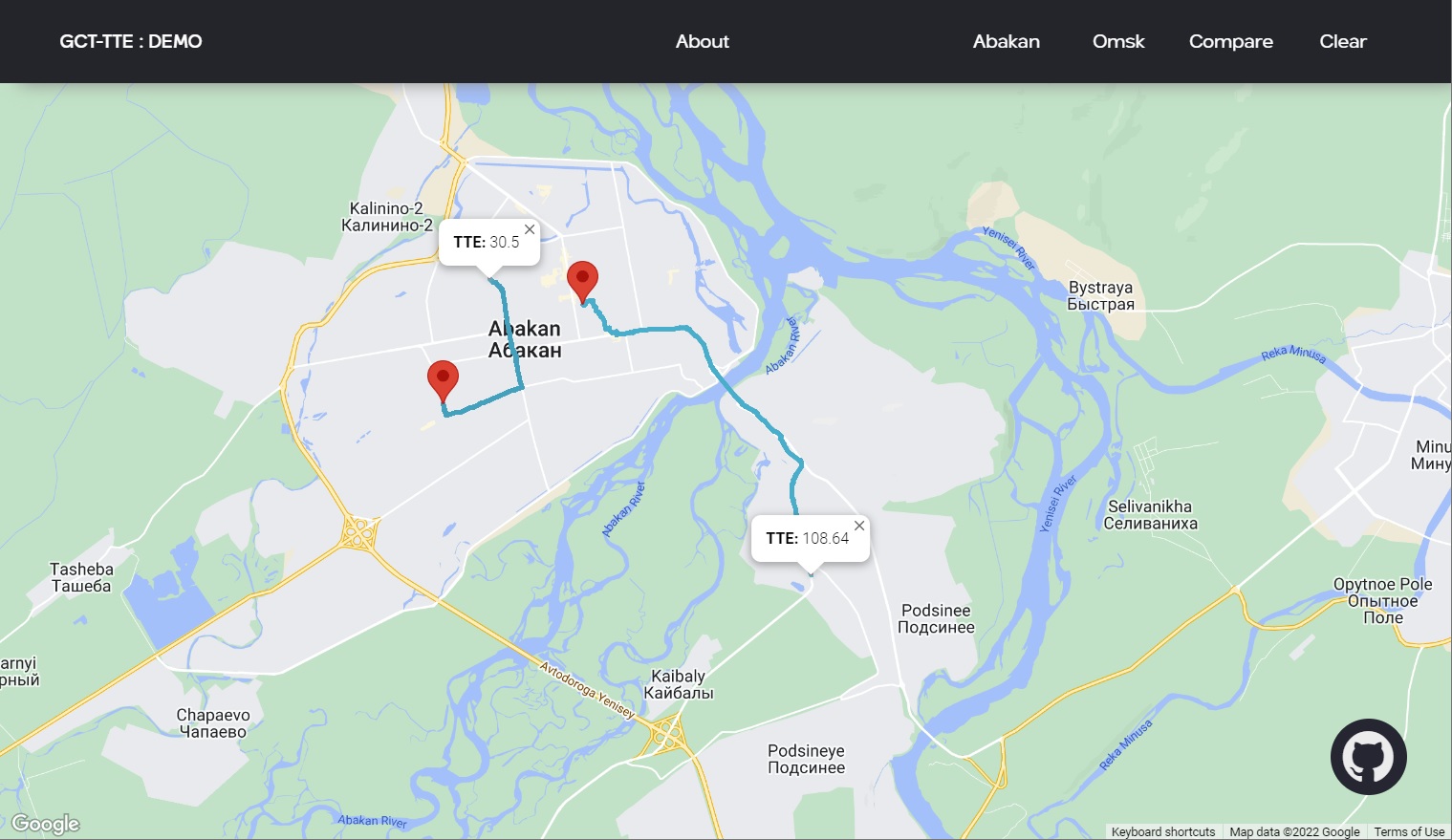}
   \caption{An interface of the demonstrational application.
   \label{demo:app}}
\end{figure*}

In order to provide access to the inference of GCT-TTE, we deployed a demonstrational application\footnote{http://gctte.online} in a website format, Figure~\ref{demo:app}. The application's interface consists of a user guide, navigation buttons, erase button, and a comparison button. A potential user can construct and evaluate an arbitrary route by clicking on the map at the desired start and end points: the system's response will contain the shortest path and the corresponding value of the estimated time of arrival. 

For additional evaluation of considered baselines, the limited number of predefined trajectories with known ground truth can also be requested. In this case, the response will contain three random trajectories from the datasets with associated predictions of WDR, DeepI2T, and GCT-TTE models along with the real travel time.  

\section*{Conclusion}
In this paper, we introduced a multimodal transformer architecture for travel time estimation and performed an extensive comparison with the other existing approaches. Obtained results allow us to conclude that the transformer-based models can be efficiently utilized as sequence encoders in the path-aware setting. Our experiments with different data modalities revealed the superior importance of graphs compared to map patches. Such an outcome can be explained by the inheritance of main features between modalities where graph data represents the same properties more explicitly. In further studies, we intend to focus on the design of a more expressive image encoder as well as consider the task of path-blind travel time estimation, which currently remains challenging for the GCT-TTE model.


\newpage

\section*{Supplementary}

\subsection*{S1 -- Features of the datasets}

\begin{table*}[htp!]

\renewcommand{\arraystretch}{1.5}

\newcolumntype{Y}{>{\centering\arraybackslash}X}

\begin{center}
\begin{tabularx}{360px}{|Y|Y|Y|}\hline
\multicolumn{1}{|Y|}{Graph node feature \cellcolor[HTML]{DDDDDD}}&
                   \multicolumn{1}{Y}{Values \cellcolor[HTML]{DDDDDD}}&
                   \multicolumn{1}{|Y|}{Description \cellcolor[HTML]{DDDDDD}}\\[2pt]
\hline
Style & undefined, archway, crosswalk, stairway, bridge, overground way, invisible, normal, park path, park footpath, subway, pedestrian bridge, underground way, tunnel, living zone, ford & additional road segments categories  \\ \hline
Road class & fake road, intra-quarter driveway, dirt road, other city street, main city street, highway, intercity road,  federal highway, cycle path, walkway & general road segments categories \\ \hline
Length & $\mathbf{Z_{+}}$ & length of a road segment in meters \\ \hline
Width & $\mathbf{Z_{+}}$ & width of a road segment in meters \\ \hline
Def  speed & \{3, 15, 20, 60, 90\} & speed limit on a road section in km/h \\ \hline
Lanes & \{0, 1, 2, 3, 4, 5\} & number of lanes in a road segment \\ \hline
Barrier & \{0, 1\} & defines the presence of road barriers \\ \hline
Payment flag & \{0, 1\} & defines a road segment as tol \\ \hline
Turn restrictions & \{0, 1\} & defines an ability to turn on a road section \\ \hline
Pedo offset & \{0, 1\} & defines the presence of crosswalk offsets \\ \hline

Bad road & \{0, 1\} &  defines the condition of a road segment \\ \hline \multicolumn{3}{c}{}\\[-0.5em] \hline

\multicolumn{1}{|Y|}{Route feature \cellcolor[HTML]{DDDDDD}}&
                   \multicolumn{1}{Y}{Values \cellcolor[HTML]{DDDDDD}}&
                   \multicolumn{1}{|Y|}{Description \cellcolor[HTML]{DDDDDD}}\\[2pt]
\hline
Nodes &$\{\hat{V} \subset V\}$ & subset of nodes \\ \hline
Dist to a & $\mathbf{Z_{+}}$ & length of a segment between actual start point and its projection on the first edge \\ \hline
Dist to b & $\mathbf{Z_{+}}$ & length of a segment between actual end point and its projection on the last edge \\ \hline
Start point part & $\mathbf{Z_{+}}$ & part of the first edge where the trip starts in meters \\ \hline
Finish point part & $\mathbf{Z_{+}}$  & part of the last edge where the trip ends in meters \\ \hline
Start UTC & $\mathbf{Z_{+}}$ & start time of the trip in UTC format \\ \hline
Real time of arrival & $\mathbf{Z_{+}}$ & trip duration in second \\ \hline
Real dist* & $\mathbf{Z_{+}}$ & Actual traveled distance in meters \\ \hline
Rebuild count* & $\mathbf{Z_{+}}$ & number of route rebuilds that corresponds to the destination change \\ \hline
\end{tabularx}
\end{center}
\end{table*}

\newpage

\subsection*{S2 -- split algorithm}

\begin{algorithm}[H]
\caption{\small{Splitting long segment}}
\begin{algorithmic}[1]
\Function{split}{$p_1$, $p_2$} \Comment{Two points [latitude, longitude]}
    \State $route\_centers = [\;]$
    \While{$\rho(p_1, p_2) \geq 200$}
        \State $a = [(p_1[0] + p_2[0]) / 2, (p_1[1] + p_2[1]) / 2]$
        \While{$\rho(x, a) \geq 200$}
            \State $a = [(p_1[0] + a[0]) / 2, (p_1[1] + a[1]) / 2]$
        \EndWhile
        \State $route\_centers \gets a$
        \State $p_1 = a$
        \EndWhile
    \State $route\_centers \gets [(p_1[0] + p_2[0]) / 2, (p_1[1] +p_2[1]) / 2]$ \Comment{Taking the middle of the final segment}
    \State return $route\_centers$
\EndFunction
\end{algorithmic}
\end{algorithm}

\subsection*{S3 -- collect algorithm}

\begin{algorithm}[H]

\caption{\small{Calculating an overall centroid of densely spaced points}}

\begin{algorithmic}[1]

\Function{collect}{$p_1$, $p_2$, $index$}
    \State $center = [0, 0]$
    \State $i = index$
    \State $k = 2$
    \While{$(i + k < len(path))$ and $(\rho(path[i], path[i+k]) < 200)$}
        \State $k = k + 1$
    \EndWhile
    \For{each $m \in [0, \ldots, k)$}
        \State $center[0] = center[0] + path[i + m][0]$
        \State $center[1] = center[1] + path[i + m][1]$
    \EndFor
    \State $center[0] = \dfrac{center[0]}{k}$
    \State $center[1] = \dfrac{center[1]}{k}$
    \State
    \State return $(center, k - 1)$ \Comment{Returns both values at once}
\EndFunction

\end{algorithmic}
\end{algorithm}

\subsection*{S4 -- Learning \& model parameters for the path-aware approaches}

\begin{table*}[!h]

    \renewcommand{\arraystretch}{1.6}

    \centering

    \newcolumntype{Y}{>{\centering\arraybackslash}X}

    \sisetup{detect-weight,mode=text}
    \renewrobustcmd{\bfseries}{\fontseries{b}\selectfont}
    \renewrobustcmd{\boldmath}{}
    \newrobustcmd{\B}{\bfseries}
    \begin{minipage}[t]{1\linewidth}
    \label{table:results_path_blind}
     \begin{tabularx}{360px}{|c|X|X|}
    \hline
       Name & Optimizer hyperparameters & Model hyperparameters \\ \hline
       DeepIST & lr = 0.000001, optimizer = Adam, scheduler = ReduceLROnPlateau, scheduler\_patience=1000 &
       G1 = 0.1, G2 = 0.1, G3 = 0.01, max\_seq\_len = 15  \\ \hline
       DeepTTE  & lr = 0.0005, optimizer = AdamW, batch\_size=4 &
       num\_filter = 32,
pooling\_method = "attention",
kernel\_size = 6,
alpha = 0.65,
num\_final\_fcs = 3,
final\_fc\_size = 128 \\ \hline
        DeepI2T  & lr = 0.01, optimizer = SGD, scheduler = ReduceLROnPlateau,
        scheduler\_regime = 'min', scheduler\_patience=2, scheduler\_factor=0.5,
        batch\_size=64 &
        without\_direction = False,
without\_conv = False,
without\_line = False,
without\_flow = False,
LSTM\_hidden\_size = 338,
bidirectional = False,
LINE\_embedding\_size = 100  \\ \hline
        WDR & lr=0.000001,
        optimizer = Adam, batch\_size=32 &
        hidden\_size = 100,
num\_layers = 1,
bidirectional=False,
deep\_input\_size = 16,
wide\_input\_size = 528 \\ \hline
        MURAT & lr = 0.01, optimizer = Adam &
        alpha = 0.2,
num\_epochs = 100 \\ \hline
    \end{tabularx}
    \end{minipage}
    \vspace{2px}

\end{table*}


\subsection*{\textbf{Declarations}}
\vspace{-12pt}

\begin{backmatter}

\section*{Ethics approval and consent to participate}
Not applicable.

\section*{Consent for publication}
Not applicable.

\section*{Availability of data and materials}
Considered models and datasets are available in the project's GitHub repository.

\section*{Competing interests}
The authors declare that they have no competing interests.

\section*{Funding}
Not applicable.

\section*{Authors contributions}
V.M., V.C., and A.I.: Software, Data curation, Validation, Visualization; 
V.P.: Software, Visualization, Conceptualization, Methodology, Writing (original draft); 
N.S.: Conceptualization, Methodology, Supervision, Writing (review \& editing).

\section*{Acknowledgements}
Authors are grateful to \anonymize{Vladislav Zamkovy} for the help with application deployment.

\bibliographystyle{bmc-mathphys} 
\bibliography{literature.bib}      

\end{backmatter}
\end{document}